\documentclass[english,11pt]{article}

\usepackage[utf8]{inputenc} 
\usepackage[T1]{fontenc}    
\usepackage{graphicx}
\usepackage{hyperref}       
\usepackage{url}  
\usepackage{amsmath,amssymb}
\usepackage{booktabs}       
\usepackage{amsfonts}       
\usepackage{nicefrac}       
\usepackage{microtype}      
\usepackage{lipsum}
\usepackage{geometry}

\author{
  Cynthia Dwork\thanks{
  Harvard John A. Paulson School of Engineering and Applied Sciences, Cambridge, MA, USA; Radcliffe Institute for Advanced Study; Microsoft Research; 
    \texttt{dwork@seas.harvard.edu}}
   \and
   Christina Ilvento\thanks{
    Harvard John A. Paulson School of Engineering and Applied Sciences, Cambridge, MA, USA; 
    \texttt{cilvento@g.harvard.edu}}
    \and
    Guy N. Rothblum\thanks{
    Department of Computer Science and Applied Mathematics,
    Weizmann Institute of Science, Rehovot, Israel;
    \texttt{rothblum@alum.mit.edu}}
    \and
    Pragya Sur\thanks{
    Center for Research on Computation and Society, Harvard John A. Paulson School of Engineering and Applied Sciences,
    Cambridge, MA, USA,
    \texttt{pragya@seas.harvard.edu}}}
\title{Abstracting Fairness: Oracles, Metrics, and Interpretability} 

\newtheorem{theorem}{Theorem}[section]
\newtheorem{assumption}[theorem]{Assumption}
\newtheorem{definition}[theorem]{Definition}
\newtheorem{lemma}[theorem]{Lemma}
\newtheorem{remark}[theorem]{Remark}
\usepackage[ruled, vlined,commentsnumbered]{algorithm2e}
\newcommand{\twopartdef}[4]
{
	\left\{
		\begin{array}{ll}
			#1 & \mbox{if } #2 \\
			#3 & \mbox{if } #4
		\end{array}
	\right.
}

\renewcommand{\dh}{d_{\mathrm{H}}}

\newcommand{\X}{{\cal X}}
\renewcommand{\P}{c}

\newcommand{\zo}{\{0,1\}}
\newcommand{\zoX}{\{0,1\}^{|\X|}}

\newcommand{\f}[1]{#1^{\mathrm{flip}}}

\newcommand{\ls}[1]{{#1}^0}
\newcommand{\rs}[1]{{#1}^1}
\newcommand{\nX}{|\X|}
\newcommand{\Ora}{{\cal O}}
\newcommand{\WOra}{\tilde{\cal O}}

\newcommand{\Gzero}[2]{G_{00}^{\,{#1} \rightarrow {#2}}}
\newcommand{\Gone}[2]{G_{01}^{\,{#1} \rightarrow {#2}}}
\newcommand{\Gdui}[2]{G_{10}^{\,{#1} \rightarrow {#2}}}
\newcommand{\Gtre}[2]{G_{11}^{\,{#1} \rightarrow {#2}}}
\newcommand{\Graph}[2]{G^{\,{#1} \rightarrow {#2}}}
\newcommand{\CSa}[2]{\mathrm{CheckSituationA}(\,{#1}, {#2} \,)}
\newcommand{\In}{\mathbf{In}}
\newcommand{\U}{\mathcal{U}}

\newcommand{\LHS}{\mathcal{LHS}}
\newcommand{\RHS}{\mathcal{RHS}}

\newcommand{\B}{\mathcal{B}}

\newcommand{\MR}{\mathrm{MergeOnes}}
\newcommand{\ML}{\mathrm{MergeZeros}}
\newcommand{\W}{\mathcal{W}}
\newcommand{\Aw}{\mathcal{A_W}}
\newcommand{\Vp}{\text{\emph{valid pair}}}

\newcommand{\T}{\mathcal{T}}
\newcommand{\bV}{\bm{V}}
\newcommand{\bVB}{\bm{V}_{\B}}
\newcommand{\bW}{\textrm{{\bf Orb}}}
\newcommand{\C}{\ell}
\newcommand{\WO}{\WOra}

\newcommand{\WOHam}{\cal O _{\cal H}}

\newcommand{\Ca}{\text{{close alignment}}}
\newcommand{\transport}[2]{{\cal C}({#1} \rightarrow {#2})}

\usepackage{bm}
\usepackage[normalem]{ulem} 

\begin{document}
\maketitle

\begin{abstract}
It is well understood that classification algorithms, for example, for deciding on loan applications, cannot be evaluated for fairness without taking context into account. 
  We examine what can be learned from a {\em fairness oracle} equipped with an underlying understanding of ``true'' fairness.  The oracle takes as input a (context, classifier) pair satisfying an arbitrary fairness definition,
  and accepts or rejects the pair according to whether the classifier satisfies the underlying fairness truth.    Our principal conceptual result is an extraction procedure that learns the underlying truth; moreover, the procedure can learn an approximation to this truth given access to a {\em weak} form of the oracle.  Since every ``truly fair'' classifier induces a coarse metric, in which those receiving the same decision are at distance zero from one another and those receiving different decisions are at distance one, this extraction process provides the basis for ensuring a rough form of {\em metric fairness}, also known as {\em individual fairness}.
  
  Our principal technical result is a higher fidelity extractor under a mild technical constraint on the weak oracle's conception of fairness.  Our framework permits the scenario in which many classifiers, with differing outcomes, may all be considered fair.

Our results have implications for {\em interpretablity} -- a highly desired but poorly defined property of classification systems that endeavors to permit a human arbiter to reject classifiers deemed to be ``unfair'' or illegitimately derived.


\end{abstract}

\section{Introduction}
Definitions of fairness, for example, in the context of accept/reject classification algorithms, mostly fall into two main categories: group fairness definitions are requirements on various forms of statistical equality in the treatment of disjoint demographic groups; {\em individual} (or {\em metric}) fairness requires that individuals that are similar with respect to the classification task at hand should be treated similarly by the classifier.  
Although intuitively appealing, group fairness definitions suffer from internal inconsistency and incompatibility~\cite{awareness,Chou17,KMR16,baer2019fairness}.
(See also~\cite{NeilWinship}.)

On the other hand, individual fairness requires a task-specific similarity metric, which may be difficult to find.\footnote{See, however, the recent proposal of \cite{Ilv19}.} 
  {\em Counterfactual Fairness}, proposed by Kusner {\it et al.} in 2017~\cite{kusner2017counterfactual}, is an approach to capturing individual fairness via {\em counterfactual reasoning} as put forth by Pearl~\cite{pearl2009causality}.
Counterfactual fairness seeks to prevent discrimination based on protected attributes, such as race or sexual preference, by requiring that individuals' outcomes ``would have been'' the same in a counterfactual world in which these attributes have different values.  To make such an assertion, the definition relies on a causal model that captures the ways in which these attributes influence other attributes relevant to classification. Thus, to evaluate whether a predictor for loan default is counterfactually fair for sexual orientation, one would construct a causal model reflecting the relationships between sexual orientation and the features weighed by the predictor, and then determine whether the predictions are inappropriately dependent on orientation. In this definition, the causal model replaces the metric as the specification of fairness.  The classifier is evaluated for fairness in the context of the causal model just as in metric fairness the classifier is evaluated for fairness in the context of the metric.
%
%

As widely noted, and partially addressed in later work (Kilbertus {\it et al.}, 2019~\cite{kilbertus2019sensitivity}), this approach suffers from the fact that different data generation models can give rise to the same distribution on outcomes. In particular, a blatantly unfair classifier can satisfy the definition when paired with a suitably contrived model, showing that the choice of model is itself a vector for unfairness. 

More generally, the maxim ``All models are wrong but some models are useful'' highlights the dangers of a fairness definition that evaluates a classifier in the context of a stated model: What are the semantics of having the (model, classifier) pair satisfy the definition when the model is wrong (which is always the case!)?  

To complete the counterfactual fairness approach (among others), one might assume the existence of an expert that can judge whether or not a classifier is ``truly fair'' in a given context. For example, a domain expert may reject a (causal model, classifier) pair for home loan decisions that satisfies the technical definition of counterfactual fairness but in which the model has been contrived to use zip code instead of race in order to obfuscate racial bias.
In this work we investigate what can be learned by interacting with such an expert.
%

We abstract the problem by instantiating ``true fairness'' (and the expert who knows what this is) via an oracle that holds a collection $\T\subseteq \zo^{|\X|}$ of vectors specifying the classification outcome for each individual in the universe $\X$ of possible individuals\footnote{We focus on the case of binary, deterministic classifiers. Such a classifier can only satisfy individual fairness if for all individuals $u,v$, $d(u,v)\in\zo$, where $d$ is the task-specific metric.  In Remark~\ref{rem:amplification}, we discuss amplification of this technique to a richer class.}. Each $t \in \T$ corresponds to a classifier that the oracle considers to be fair, at least in some context. As an example, one might imagine that the oracle has access to the true data generation model, and it evaluates classifiers in this single context.


Our principal conceptual result is an (inefficient) {\em extraction procedure} that learns the underlying truth (collection $\T$) held by the oracle under the assumption that the contexts of interest are of bounded size.
Once the assumption is cleanly stated it is not surprising that $\T$ can be extracted by brute force, so this first contribution is the conceptual framing of the problem (Sections~\ref{sec:definitions} and~\ref{sec:maincont}). 
This result makes no assumptions about the set of fair classifiers accepted by the oracle, nor about the particular context(s) that make the oracle accept a classifier. We extract the full set of classifiers for which there exists {\em some} context that makes the oracle accept.

Under the assumption that counterfactual fairness (or any other causality-based definition, such as path-specific effects \cite{nabi2018fair}), combined with the true causal model (or an appropriate approximation), genuinely captures fairness, our results imply that one can extract, from an oracle with access to the true model, a coarse metric for individual fairness. This holds because every classifier induces a coarse metric in which those receiving positive decisions are at distance zero from one another, and similarly for those receiving negative decisions. (See Remark~\ref{rem:amplification}.)

We then turn to {\em weak} oracles, which solve a more relaxed promise problem.  Each weak oracle $\WOra$ is a relaxation, based on a given notion of closeness of classifiers, of a strong oracle, $\Ora$. Weak oracles always accept the (context, classifier) pairs accepted by their strong counterparts, but only reject (context, classifier) pairs where the classifier is ``far'' from an accepted classifier for the given distance notion.  


We consider two types of closeness in defining weak oracles: Hamming distance, where the reconstruction problem is straightforward, provided the members of $\T$ are sufficiently separated\footnote{Much as it is possible to learn a mixture of Gaussians provided the means are sufficiently far apart.}, 
and an asymmetric {\em transportation cost ${\cal C}$} that does not satisfy the triangle inequality. Our transportation cost is closely related to individual fairness: $\transport{t}{c}$ captures the number of pairs of individuals that are treated similarly in $t$ but differently in $c$.
In essence, the transportation cost notion requires {\em less} of the oracle: $\delta$-weak oracles\footnote{Oracles only guaranteed to reject classifiers at distance greater than $\delta$ from all $t \in \T$.} with this notion of distance may not know {\em how} individuals should be treated for the task at hand, but may have a sense of who should be treated similarly to whom.  This lack of decisiveness on the part of the oracle makes extraction much more difficult.  Not only does it lead to a transportation cost that is not even a distance function, but it also limits what can possibly be extracted even if $\delta = 0$: under this notion, the distance between a classifier and its complement is 0!
\footnote{Our techniques apply to a symmetrized version of $\transport{t}{c}$, defined by the fraction of pairs of individuals that disagree between $t$ and $c$ (Section \ref{sec:maincont}). This case is, in fact,  easier than the transportation cost.} In consequence, rather than aiming to extract the set of fair classifiers, we extract a set of fair partitions, where each partition specifies which individuals are similar to each other. The partition can also be viewed as a coarse metric.
%
%

 Our principal technical result is a high fidelity extractor in the transportation cost model, under a mild technical constraint on the weak oracle's conception of fairness. 
For $t \in \T$, define $\f{t}(x) = 1-t(x)$ for all $x \in \X$.  The assumption is:
for $t \in \T$ for which the weak oracle rejects $\f{t}$, it also rejects all classifiers very close (in Hamming distance) to $\f{t}$.



\paragraph*{Interpretability} Our results have implications for {\em interpretablity} -- a highly desired but poorly defined property of classification systems that endeavors to permit a human arbiter to reject classifiers deemed to be ``unfair'' or illegitimately derived.  If ``interpretability'' permits a knowledgeable human to distinguish truly fair from truly unfair classifiers, then there is a procedure to extract from the human information a measure of similarity for pairs of individuals. Roughly speaking, we can get our hands on a metric, even when the closeness notion for classifiers is the Hamming distance on their vector representation, which is unrelated to metric fairness! 


\begin{remark}
\label{rem:amplification}
In this work, the ``metric'' we extract from the oracle is crude: all distances are either 0 or~1. Metrics of this type can be amplified to yield a richer class of metrics by considering a collection of oracles with varying tolerance for unfairness. For example, given a metric $d: \X \times \X \rightarrow [0,1]$, we can instantiate $k$ approximations $\{d'_1,\ldots, d'_k\}$ of $d$ such that $d'_i(u,v) := 1$ if $d(u,v) > \frac{k}{i}$ and $0$ otherwise. Given access to an oracle for each threshold, we can apply the extraction procedure multiple times to (approximately) recover this set of $\{0,1\}-$metrics. The recovered collection can then be combined to form an approximation of $d$, using the threshold combination procedure developed in~\cite{Ilvento19}. See also \cite{gillen2018online, JKMR16,jung2019eliciting} for demonstrations of the usefulness of coarse metrics.
\end{remark}

\paragraph*{Related Work}
There is a vast literature on algorithmic fairness.  The theory of algorithmic fairness was first studied by Dwork {\it et al.} in 2012~\cite{awareness}. In addition to defining individual fairness, this work noted that sensitive attributes may be holographically embedded in the data, showed the benefits of utilizing, rather than trying to suppress, the sensitive information; showed the power of Individual Fairness when given a metric; examined the group fairness property of demographic parity and gave examples motivating its dismissal as a fairness solution concept, and provided a metric-based approach to Fair Affirmative Action.
Earlier work suggested concrete approaches based on training on a modified dataset in which the proportion of positive labels is equal in disjoint demographic groups, in the hopes that a classifier trained on these new labels will imbibe the group fairness properties of the training data~\cite{pedreshi2008discrimination,kamiran2009classifying}. A second approach added a regularization term to the classification training objective to quantify the degree of bias or discrimination~\cite{kamishima2011fairness,calders2010three}. 
Subsequent work saw heavy investment in algorithms satisfying group-based criteria, even in the face of the negative results about the compatibility of natural group fairness objectives~\cite{NeilWinship,Chou17,KMR16,gerry,DI2018}.
Individual fairness, predicated on access to a similarity metric, proceeded more slowly, although the literature contains several works extending the theory~\cite{DI2018,RothblumYona,gillen2018online,kim2018fairness}.  Recent work \cite{Ilv19} combines insights from HCI and computational learning theory to learn an approximation to a metric known to a human arbiter with surprisingly few queries.
An intriguing ``middle ground'' enforces {\em calibration} (in the case of scoring functions~\cite{HKRR18}) {\em simultaneously} for large numbers of intersecting subpopulations (see~\cite{DKRRY19} for a treatment of fair {\em rankings} in this setting).  A variant of the multiple intersecting groups approach~\cite{gerry} enforces {\em Equalized Odds}~\cite{hardt2016equality} among all pairs of groups simultaneously.  An economics justification for Equalized Odds is put forth in~\cite{endogenous2020}. Equalized Odds and related candidate fairness criteria are criticized through the lens of graphical models~\cite{baer2019fairness}.

Still other work employs deep learning to build {\em fair representations} of individuals that, speaking intuitively, retain much useful information for classification or even transfer learning, but ``screen out'' sensitive demographic information~\cite{Zem13,edwards2015censoring,creager2018learning}.
Finally, there is also a vast literature on {\em interpretability}.  See~\cite{lipton2018mythos} for a discussion of what this might mean (and hurdles to be overcome); the course notes of Lakkaraju~\cite{LakkarajuCourse2019} contain a wealth of examples and references for this literature.

Our work was inspired by the elegant proposal of Counterfactual Fairness by Kusner, Loftus, Russell, and Silva~\cite{kusner2017counterfactual}.  A related definition of fairness concentrates on path-specific effects~\cite{nabi2018fair} (see also~\cite{kilbertus2017avoiding}).  
Kilbertus {\it et al.} design tools to assess the sensitivity of fairness measures to unmeasured confounding for a popular class of noise models~\cite{kilbertus2019sensitivity}.

\paragraph*{Organization} The rest of this paper is organized as follows.  Section~\ref{sec:definitions} introduces the definitions used in this work.
Section~\ref{sec:maincont} states the main contributions and motivates 
our use of oracles. Section \ref{sec:algorithms} describes our algorithms, and Section \ref{sec:assumptions} concludes with a discussion of necessary assumptions. 

\section{Definitions}\label{sec:definitions}

We consider a universe $\X$ of individuals, each represented by a vector of $p$ attributes.
We will assume each vector of attributes represents a unique individual.
Determining whether or not the representation of the individuals is sufficient to permit fair classification is a fascinating topic beyond the reach of this paper; here we assume an affirmative answer.  Since our work may be viewed as negative results, this assumption only strengthens the contribution.

A {\em classifier} maps individuals to $\zo$, $C:\X\rightarrow \zo$.  It is often convenient to think of classifiers as vectors $c \in \zo^{|\X|}$, with $c_i\in\zo$ being the classification of the $i$th individual in some canonical ordering.  We completely identify an individual and its index $i$, so we will often write $i \in \X$ to denote the $i$th individual in this ordering.

It is sometimes convenient to think of a classifier as partitioning $\X$ into two groups according to their classification outcomes.
Unless otherwise specified, we use lower case letters to denote classifiers and the corresponding upper case letter to denote the partition.
For a classifier $c$, we let $\ls{c} = \{i \in \X | c_i=0\}$ and $\rs{c} = \{i \in \X | c_i = 1\}$. 
 We sometimes refer to $\ls{c}$ as the {\em Left Hand Side} of the partition $c$, denoted $\LHS(c)$, and $\rs{c}$ as the {\em Right Hand Side}, denoted $\RHS(c)$. 
 The {\em flip} of a partition is a swap of its left and right sides; in vector form, $\f{c} = 1-c$, {\it i.e.}, $\forall i \in \X, \f{c}_i=1-c_i$.
  Constant classifiers have the property that for some $v \in \zo$, $c_i = v,~ \forall i \in \X$. 
 
\begin{figure}[h]
\centering 
\begin{minipage}{.49\textwidth} 
\centering 
\includegraphics[width=0.6\linewidth]{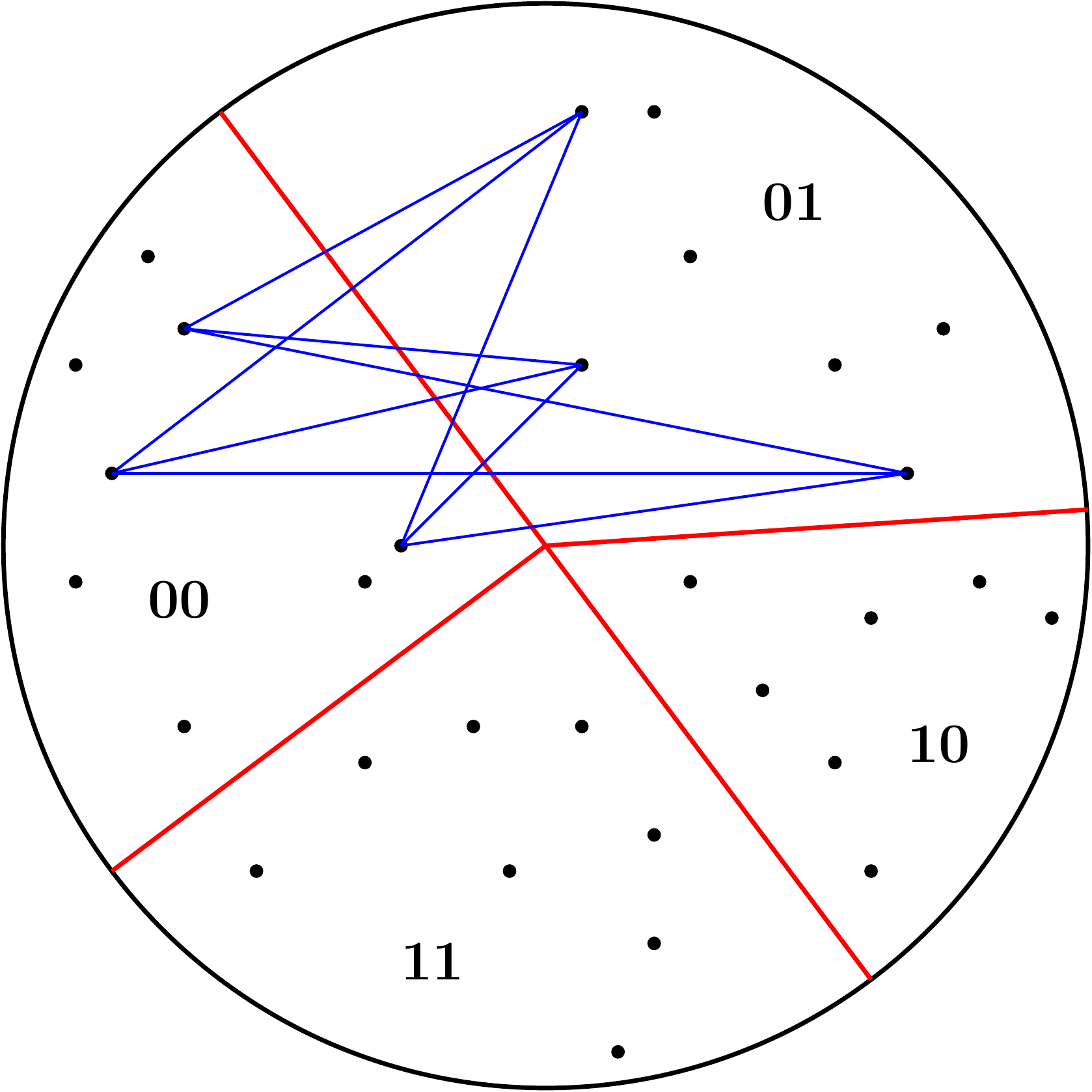}
\end{minipage}
\begin{minipage}{.49\textwidth} 
\centering 
\includegraphics[width=0.6\linewidth]{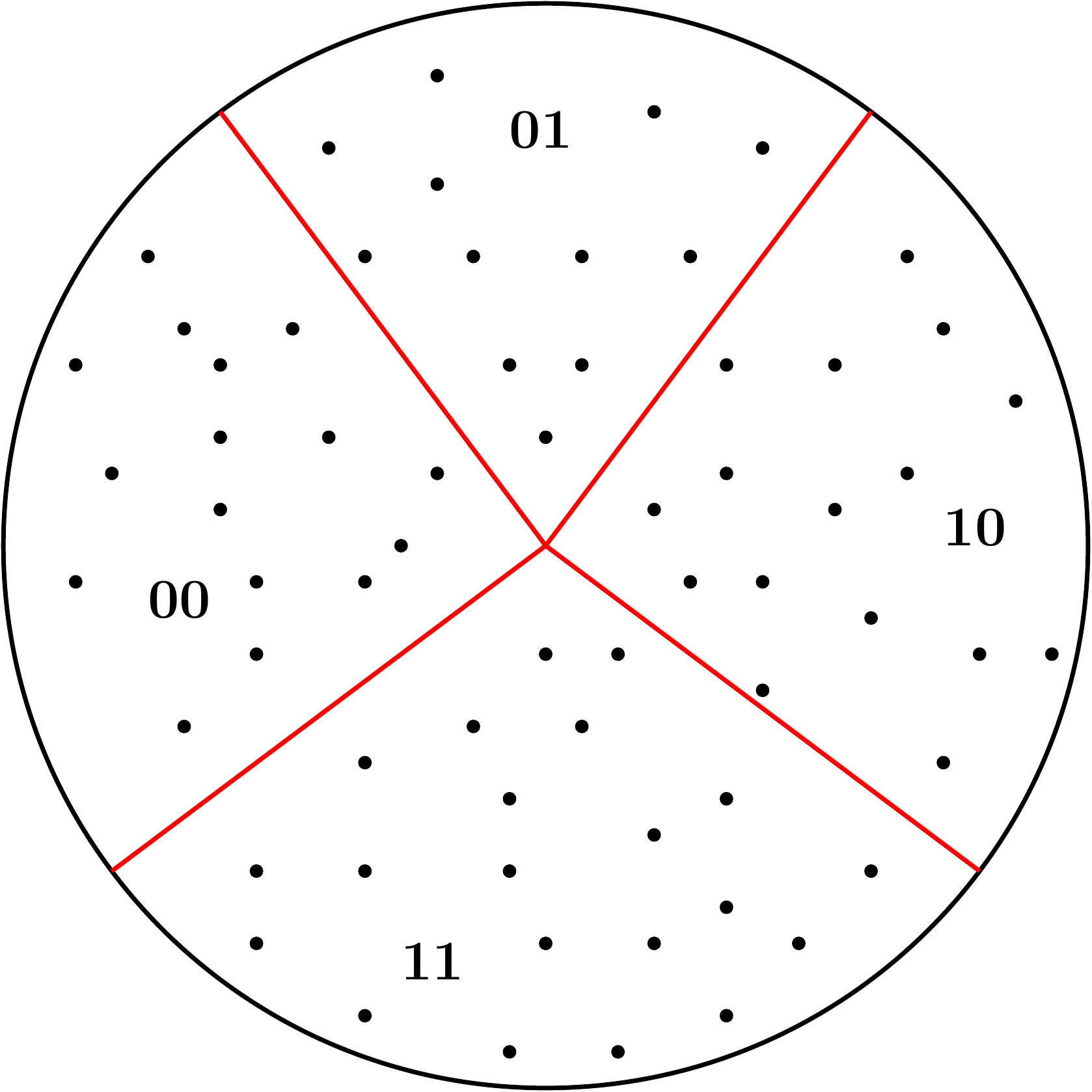}
\end{minipage}
\caption{ (Left)
The (complete) labeled graph $\Graph{u}{v}$ for an ordered pair of classifiers $(u,v)$.  There is a vertex for each $i \in \X$. Vertices are labeled with two-bit strings indicating their classifications under $u$ (first bit) and $v$ (second bit).  So vertices $i$ in the quadrant $\Gzero{u}{v}$ have $u_i=v_i=0$, vertices in $\Gone{u}{v}$ have $u_i = 0$ and $v_i = 1$, and so on. The transportation cost $\transport{u}{v}$ is captured by the number $|\Gzero{u}{v}|\cdot|\Gone{u}{v}|$ of edges between the 00 and 01 quadrants (a few drawn in blue on left) plus the number of edges between the 11 and 10 quadrants (none drawn). For example, if $i \in \Gzero{u}{v}$ and
$j \in \Gone{u}{v}$ this says that the edge $(i,j)$ was monochromatic (both vertices colored zero) in $u$ but is polychromatic in $v$ (because $v_i = 0 \not= v_j = 1)$.  (Right) Illustration of Assumption~\ref{assm:summary}(2): all quadrants are substantial.
}
 \label{fig:faithfulness}
\end{figure}

It is also sometimes convenient to think of individuals in $\X$ as vertices, and to think of the classifier as a two-coloring of the complete graph on~$\X$ (see Figure \ref{fig:faithfulness}). Monochromatic edges indicate pairs of individuals who are treated the same by the classifier.

\paragraph*{Contexts and Valid Pairs}  
Many fairness notions require that classifiers be considered in some form of {\em context}.  For example, in the case of counterfactual fairness the context is given by a causal model%
\footnote{See~\cite{baer2019fairness} for a general discussion of the need for context.}.
We therefore abstract the notion of a fairness definition  
$\W$ as a set of (context, classifier) pairs.  

\begin{definition}[validity]
If $(\Aw,c) \in \W$ then $(\Aw,c)$ is said to be a {\em \Vp{} under $\W$}.  In our work $\W$ is typically fixed, in which case we may simply refer to {\em valid pairs}. 
\end{definition}


\paragraph*{Boundedness}
We assume there is a procedure for enumerating all contexts, whose running time is a fixed function of $|\X|$. For example, we might consider the case in which the context is given by a causal graph constrained to have a number of vertices linear in~$p$ (the number of attributes)  and the functions computed at each vertex can be described by circuits of size polynomial in~$|\X|$. We note that without this assumption it is not even clear how to represent a context $\Aw$ for the purposes of determining whether or not some $(\Aw,c) \in \W$.

\paragraph*{Oracles}
We view the fairness definition $\W$ as a filter, and hypothesize the existence of an {\em oracle} to rule on the acceptability of valid pairs. 
A useful intuition, for example, with counterfactual fairness in mind, is that the oracle knows the true data generation model $\Aw^*$, and is willing to accept exactly valid pairs $(\Aw^*,c) \in \W$; alternatively, the oracle may be willing to accept valid pairs $(\Aw,c)$ whenever $\Aw$ enjoys certain properties.  
However, we make no explicit assumptions: Formally, the oracle is specified by a subset of~$\W$. It takes as input a valid pair $(\Aw,t) \in \W$ and either accepts ($\Ora(\Aw,t)=1$) or rejects ($\Ora(\Aw,t)=0$). 

\begin{definition}[Strong Oracle]
   A strong oracle is completely specified by the valid pairs that it accepts.
\end{definition}
\noindent It is convenient to name the collection of classifiers associated with acceptance by the strong oracle, that is, to define $\T = \{t \in \zo^{|X|} \,|\, \exists \Aw:~  \Ora(\Aw,t) = 1\}$. 

\paragraph*{Weak Oracles}
Every weak oracle is a relaxation of a strong oracle.
Weak oracles differ from their corresponding strong oracles by relaxation of the conditions for acceptance: weak oracles will accept whatever the associated strong oracles accept, but may also accept valid pairs.

\begin{definition}[$\delta$-Weak Oracle for Hamming distance]
\label{def:woham}
Fix an arbitrary strong oracle $\Ora$ with associated classifiers~$\T$.  For $\delta>0$ we say that $\WOHam$ is a {\em $\delta$-weak oracle relaxation of~$\Ora$}, based on the Hamming distance, if 
\begin{enumerate}
\item $\WOHam$ accepts all valid pairs accepted by~$\Ora$; 
\item $\WOHam$ rejects valid pairs whose classifiers are far (in Hamming distance) from all classifiers in~$\T$: Let $(\Aw,c) \in \W$.  If $\forall t \in \T, d_{\mathrm{H}} (c,t) > \delta$, then $\WOra(\Aw,c) = 0$. Here, for $u,v \in \zoX$, $d_{\mathrm{H}}(u,v) := \{i \in \X \ | \ u_i \neq v_i \}$.
\end{enumerate}
On the remaining valid pairs, $\WOHam$ may behave arbitrarily.
\end{definition}

The definition of a weak oracle {\em based on transportation cost} requires one additional concept.


\begin{definition}[{$\delta$-faithfulness}]
For $c,t \in \zo^{|\X|}$,  we say that $c$ is {\em $\delta$-faithful} to $t$ if
\begin{equation}\label{eq:faithful}
\frac{1}{{n \choose 2}} 
\sum_{i \neq j} \bm{1}\{t_i=t_j, c_i \not= c_j  \}  \leq \delta. 
\end{equation}
\end{definition}

Note that faithfulness
is not symmetric.  Typically, we will consider faithfulness when $c$ is a candidate classifier and $t$ is an element of the set $\T$ associated with an oracle.  
$\delta$-faithfulness suggests a natural transportation cost capturing the answer to the question, ``Starting from $t$, how many monochromatic edges in $t$ do we need to ``break'' when we transition to~$c$?''  We let $\transport{t}{c}$ denote this transportation cost  (Figure \ref{fig:faithfulness}).  This transportation cost is asymmetric and does not satisfy the triangle inequality.

\begin{definition}[$\delta$-neighborhood]
The {\em $\delta$-neighborhood} of a classifier $t \in \zo^{|\X|}$, denoted $\Gamma_\delta(t)$, is the set of all $c \in \zoX$ such that $c$ is $\delta$-faithful to $t$.
\end{definition}


\begin{definition}[$\delta$-Weak Oracle for transportation cost]
\label{def:wo}
Fix an arbitrary strong oracle $\Ora$ with associated classifiers~$\T$.  For $\delta>0$ we say that $\WOra$ is a {\em $\delta$-weak oracle relaxation of~$\Ora$}, based on the transportation cost $\cal C$, if 
\begin{enumerate}
\item $\WOra$ accepts all valid pairs accepted by~$\Ora$; $\forall (\Aw,c)$ such that $\Ora(\Aw,c)=1$, $\WOra(\Aw,c)=1$; 
\item $\WOra$ rejects valid pairs whose classifiers are far (in transportation cost) from all classifiers in~$\T$: Let $(\Aw,c) \in \W$.  If $\forall t \in \T, c \notin \Gamma_\delta(t)$, then $\WOra(\Aw,c) = 0$.
\end{enumerate}
\end{definition}
There are no further constraints on oracles other than being deterministic.  Note that there may be many weak oracle relaxations of a given strong oracle~$\Ora$.  

\section{Main Contributions}\label{sec:maincont}

Our principle contributions are extraction procedures that recover the underlying truth held by oracles. Recall that a strong oracle is associated with a set $\T$ of classifiers. Formally, an {\em extraction procedure} is a program that, using only access to an oracle $\Ora$, outputs the list $\T$ associated with~$\Ora$. Intuitively, one may imagine that this set arises from some ground truth provided by the concept $\mathcal{W}$. To illustrate, let $\mathcal{W}$ be the notion of counterfactual fairness and $\mathcal{M}$ the true causal model explaining actual functional relationships between all the relevant variables for the task. An oracle may believe that all classifiers 
that satisfy the counterfactual fairness definition with respect to this ``true'' causal model are indeed truly fair. Then, $\T$ equals the set of all  counterfactually fair classifiers with respect to $\mathcal{M}$. The data analysts have no knowledge of $\mathcal{M}$ whatsoever, but hope to learn about fair classifiers by interacting with the oracle. We provide algorithms that achieve this goal---starting with the simpler case of the strong oracle, subsequently moving on to weak oracles. 

Recall that every strong oracle $\Ora$ has an associated set $\T$ of classifiers, such that $\forall t \in \T, \, \exists \Aw$ with $(\Aw,t) \in \W$ and $\Ora(\Aw,t)=1$; $\Ora$ rejects all valid pairs $(\Aw,c)$ with $c \notin \T$. To begin with, we establish the following. 

\begin{theorem} \label{thm:strong}
For any fairness notion $\W$ satisfying the boundedness condition, and for any strong oracle $\Ora$ accepting a subset of~$\W$, there exists an extraction procedure interacting with $\Ora$ whose running time is bounded by a function of $|\X|$. The output of the extraction procedure is the set $\T$ associated with $\Ora$.\end{theorem}

Under the assumption of bounded length contexts, Theorem \ref{thm:strong} can be achieved simply via exhaustive search, since our extraction procedures are allowed to be inefficient. The primary contribution of this result is thus conceptual---that it is feasible to extract the ground truth from $\Ora$ under our framing of the problem. In the spirit of prior examples, if $\W$ is counterfactual fairness and  $\T$ the set of all counterfactually fair classifiers with respect to the true causal model (which we do not have any access to, but let's say the oracle has complete knowledge about), then in principle one can learn all of these classifiers.  Note that each of these is equivalent to a partitioning of the universe, and can therefore be viewed as a metric (albeit a simple one). Intuitively, one can interpret the oracle as a highly knowledgeable human expert with a deep understanding of the true underlying relationships between the variables relevant for the task, but who is unable to enunciate them---however, the expert is able to tell whether a classifier is fair or not by ``looking'' at it.  Our result demonstrates that, given access to such an expert, a  systematic strategy can successfully learn all the fair classifiers. 
Thus, in settings where learning the true causal model is extremely hard (if not impossible), and hence, reliably implementing counterfactual fairness (or any other causality-based notion) may be out of scope, our results suggest that developing efficient query models to interact with human experts suffices for fair classification, since these directly learn metric information from the expert (recall Remark 1). 

While it is helpful to think of an all-knowing expert, who can accurately identify fair classifiers and task-appropriate contexts, our framework can be applied to {\em any} expert. We can also extract from an imperfect expert, {\textit{e.g.}}~one who can only reason about simple contexts, and accepts a subset of the fair classifiers (or even accepts some unfair ones!). The better the expert, the better the classifiers (or metrics) we extract will be.

%

\paragraph*{Weak Oracles}



A weak oracle accepts every valid pair accepted by a strong oracle; in addition, it rejects valid pairs whose classifiers are far from all classifiers in $\T$. However, a weak oracle may behave arbitrarily on the remaining pairs. 
Nonetheless, we are able to extract even when we do not know how the oracle will behave on these remaining pairs, and this is a strength of our framework. Since the oracle only provides fuzzy information, in the sense that it may behave at will on several pairs, we can only hope to recover $\T$ up to some error. 
Different notions of distance that determine what should be judged ``far'' lead to different instantiations of the weak oracle. The conversation around the right notion of distance lies beyond the scope of this paper. Here, we consider two notions,  Hamming distance and transportation cost, and provide algorithms 
in each case that approximately recover $\T$. 

\paragraph*{Extraction from a Hamming distance based weak oracle}
To begin with, we consider a natural distance measure---the Hamming distance. Recall that any weak oracle is a relaxation of a strong oracle $\Ora$ with an associated set $\T$ of classifiers. A weak oracle $\WOra$ based on the Hamming distance accepts any valid pair  accepted by~$\Ora$, and rejects a valid pair $(\Aw,c)$ for which  $d_{\mathrm{H}}(c,t) > \delta$ for every $t \in \T$, and otherwise behaves arbitrarily. Recovery of an approximation to the elements in $\T$ is then trivial (via exhaustive search) as long as any two elements $t,u \in \T$ satisfy  $\dh(t,u) > 4 \delta$.
 (Details omitted.) 
 



\paragraph*{Extraction from a transportation cost based weak oracle}
Our primary technical contribution is an extraction algorithm that approximately recovers elements of $\T$ from a weak oracle based on the transportation cost $\mathcal{C}(t \rightarrow c)$ (Definition \ref{def:wo}). Theorem \ref{thm:sharpthmsumm}, stated next, says that the Sharp Extraction Algorithm (Algorithm \ref{algo:sharp}, Section \ref{sec:algorithms}) produces a list of classifiers, each of which corresponds  to a unique member of $\T$. Recall that for any classifier $c$,  $\ls{c} = \{i \in \X | c_i=0\}$ and $\rs{c} = \{i \in \X | c_i = 1\}$.


\begin{theorem}\label{thm:sharpthmsumm}
Suppose $\Ora$ is a strong oracle with an associated set $\T$ of classifiers, and $\WOra$ is a $\delta$-weak relaxation of $\Ora$ under the transportation cost $\mathcal{C}(t \rightarrow c)$ (Definition \ref{def:wo}).  
Then under Assumption \ref{assm:summary}, the list of classifiers ($P_1,\ldots,P_m,Q_1,\ldots,Q_m$) obtained from Sharp Extraction (Algorithm \ref{algo:sharp}, Section \ref{sec:algorithms}) satisfies the following: Fix any index  $j$.  There exists $t \in \T$ such that 

\begin{align}\label{eq:rel1}
\ls{P_j} & \subset \ls{t} \ \  (\text{or} \ \  \rs{t}) \qquad \text{and} \qquad
 \rs{Q_j}  \subset \rs{t} \ \  (\text{resp.} \  \ls{t}),
\end{align}
simultaneously,
\begin{align}
|\rs{P_j} \cap \ls{t}| \ \  (\text{resp.~} \  \rs{t}) & \leq \left(\frac{\tau_j - \sqrt{\tau_j^2 - 2\delta}}{2}\right) n, \quad \text{and} \nonumber\\
|\ls{Q_j} \cap \rs{t}| \ \  (\text{resp.~} \  \ls{t}) & \leq \left(\frac{\tilde{\tau}_j - \sqrt{\tilde{\tau}_j^2 - 2\delta}}{2}\right) n.
\label{eq:errors1}
\end{align}
Above, $\tau_j= |\ls{t}|/n$ and $\tilde{\tau_j} = 1-\tau_j$. 
For classifiers $P_j, Q_j$ and $ P_k, Q_k$ with different indices, the corresponding elements of $\T$ that satisfy the aforementioned property are also different.
\end{theorem}

 Sharp Extraction precisely pins down the elements of $\T$ up to a small error margin as specified by \eqref{eq:errors1}. The smaller the value of $\delta$, the lower the  overall error. In general, for every $t \in \T$, Sharp Extraction recovers nearly as many members of $t^0$ and $t^1$ as possible (without recovering the exact classification outcomes).
  To see this, observe that the fraction of pairs of individuals that $P_j$ erroneously splits in two  groups, when they belong to the same group in the underlying element of $\T$, can be bounded by 
\[\left(\frac{\tau_j - \sqrt{\tau_j^2 - 2\delta}}{2}\right) \left(\frac{\tau_j + \sqrt{\tau_j^2 - 2\delta}}{2}\right) + \left(\frac{\tilde{\tau}_j - \sqrt{\tilde{\tau}_j^2 - 2\delta}}{2}\right)\left(\frac{\tilde{\tau}_j - \sqrt{\tilde{\tau}_j^2 + 2\delta}}{2}\right) = \delta. \]
Since  $\WOra$ is a $\delta$-weak relaxation of $\Ora$, 
intuitively, one cannot hope to accurately cluster additional pairs of individuals from this weak oracle model, suggesting that Sharp Extraction achieves the best we may hope for in such a setting.

\paragraph*{Extraction from a weak oracle based on a symmetrized transportation cost} 

Extraction algorithms can also be developed for weak oracles based on the symmetrized version of the transportation cost
\begin{equation}\label{eq:stc}
 \mathcal{C}^s(u \leftrightarrow v) := \frac{1}{{ n \choose 2}} \sum_{i \neq j} \left[ \boldsymbol{1} \{u_i = u_j, v_i \neq v_j \} + \boldsymbol{1}\{u_i \neq u_j, v_i = v_j \} \right].    
\end{equation}
A weak oracle $\WOra$ based on this notion accepts any valid pair accepted by its associated strong oracle~$\Ora$, and rejects any valid pair $(\Aw,c)$ for which  $\mathcal{C}^s(t \leftrightarrow c) > \delta$ for every $t \in \T$. This case is, in fact, easier to handle than the asymmetric version. Thus, algorithms that work for weak oracles based on the asymmetric transportation cost
can be simplified to suit the needs of a weak oracle based on the symmetrized version \eqref{eq:stc}. 


\paragraph*{Conclusion} 

Classification algorithms cannot be evaluated for fairness without taking context into account.  Several works in the fairness literature posit the existence of fair and wise human judges, and the ability of humans to recognize unfairness when they see it seems to be a linchpin of interpretability.  
We have explored what can be learned from a fairness oracle that evaluates (context, classifier) pairs satisfying a definition of fairness, accepting or rejecting according to a hypothesized fairness ``truth''.  The oracle abstraction captures any human judge, or algorithm, or benchmark test; the extraction procedures described here do not need to ``understand'' the oracle's decisions. Even so, the procedures produce rudimentary metrics for the classification task at hand.  The procedure can be amplified to improve the expressive power of the metric. These existence proofs are evidence for the conjecture that a metric is {\em always} at the heart of fairness.

  Metrics can be combined with arbitrary loss functions to obtain individually fair classifiers satisfying a wide range of objectives~\cite{awareness}.  Metrics learned on a sample of the population can sometimes be generalized to unseen examples~\cite{Ilvento19}.  An {\em efficient} metric extraction procedure would mean that it is essentially no harder to find a metric than to build good causal models and accompanying classifiers.  This is an exciting direction for future research.
  
  \section{Extraction Algorithms}\label{sec:algorithms}

This section summarizes our extraction algorithms and key ingredients used therein.

\begin{definition}[$\delta$-Balanced classifier]
\label{def:balanced partition}
A classifier $c$ is said to be {\em$\delta$-balanced} if both 
$|\ls{c}| > \sqrt{2 \delta} \nX $ and $|\rs{c}| > \sqrt{2 \delta} \nX$.
\end{definition}

We let $\B$ denote the set of all $\delta$-balanced classifiers of~$\X$; henceforth, we simply call these the balanced classifiers.

Two classifiers are said to be aligned if they have relatively few disagreements.  Recall that, for a classifier $c \in \zoX$,
 $\ls{c} = \{ i \, | \, c_i = 0\}$ and $\rs{c}$ is defined analogously.
\begin{definition}[Close alignment]
\label{def:ca}
Classifiers $p$ and $q$ in $\zoX$ are {\em in close alignment} if
$|\ls{p} \cap \rs{q}|, |\ls{q} \cap \rs{p}|| \le\sqrt{\frac{\delta}{2}}|\X|$.
\end{definition}

Furthermore, define the following sets (Figure \ref{fig:faithfulness}) for any $v, c\in \zoX $,
    \begin{align}\label{eq:foursets}
    \Gzero{v}{c}=\{i\,|\,v_{i}=0, c_{i} =0 \}, \qquad &
    \Gone{v}{c} = \{i\,|\,v_{i}=0, c_{i} =1 \}, \\
    \Gdui{v}{c} =\{i\,|\,v_{i}=1, c_{i} =0 \}, \qquad &
    \Gtre{v}{c} = \{i\,|\,v_{i}=1, c_{i} =1 \}.
       \end{align}
The proof of Theorem~\ref{thm:sharpthmsumm} and the description of the algorithms require the following lemma.
\begin{lemma}\label{lemma:distinguish}
Let $u,v$ be arbitrary balanced classifiers accepted by the weak oracle $\tilde{O}$ in Theorem \ref{thm:sharpthmsumm}.  Then exactly one of the following must hold:
\begin{enumerate}
\item There exists $ t \in \T$ such that $u,v \in \Gamma_{\delta}(t)$. In this case, 
\[
    \min \{|\Gzero{u}{v}|, |\Gone{u}{v}| \} \leq \sqrt{2 \delta} n \leq \min \{|\Gdui{u}{v}|, \Gtre{u}{v} \}. \qquad \text{\emph{Situation A}}
\]
\item There does not exist any $ t \in \T$ such that $u,v \in \Gamma_{\delta}(t)$. In this case, there must exist $t_1, t_2 \in \T$ such that $t_1 \neq \f{t_2}$, $u \in \Gamma_\delta(t_1), v \in \Gamma_\delta(t_2)$, and that at least one of the following holds 
\[
 \min \left\{|\Gzero{u}{v}|, |\Gone{u}{v}| \right\} > \sqrt{2\delta} n, \qquad \min \left\{|\Gdui{u}{v}|, \Gtre{u}{v} \right\}   > \sqrt{2 \delta} n.  \qquad \text{\emph{Situation B}}
\]
\end{enumerate}
\end{lemma}

Our main algorithm, Sharp Extraction, builds on Fuzzy Extraction, presented in Algorithm \ref{algo:fuzzy}.

\paragraph*{Informal Description of Fuzzy Extraction Algorithm} 
The fuzzy extraction algorithm seeks to associate candidate classifiers $c$ with elements of $\T$, roughly, guided by the transportation cost $\transport{t}{c}$. 
In particular, the algorithm aims to recover $\Gamma_\delta(t)$ for each $t \in \T$. 
As with the reconstruction from a strong oracle, by the boundedness requirement for contexts, we can find 
$\bV = \{c\in \zoX \, \mid \, \exists \Aw : ~ \WOra(\Aw,c)=1\}$ by enumeration.  
The algorithm starts by finding $\bV$. 
The algorithm then prunes out all unbalanced classifiers, setting $\bVB= \bV \cap \B$. This is the starting point for recovering~$\T$, the classifiers associated with the strong oracle $\Ora$ of which $\WOra$ is a relaxation.

At a high level, Fuzzy Extraction works as follow: for an arbitrary pair $u,v \in \bVB$, the algorithm checks whether $u$ and $v$ are both in $\Gamma_\delta(t)$ for some $t \in \T$. From Lemma \ref{lemma:distinguish}, this can be detected simply by inspection, that Situation~A holds.
If so, the algorithm clusters $u$ and $v$ into the same group, and otherwise, to different groups. In this manner, the algorithm builds up a collection of sets, each of which contains classifiers that are all in $\Gamma_\delta(t)$ for some $t \in \T$. 

In addition, the algorithm takes special care to track when $u,v \in \bVB$ are in \Ca{} (Definition~\ref{def:ca}; roughly speaking, they are closely aligned if they are close in Hamming distance). Using this information, the algorithm builds a collection of sets, which we call {\em orbits}, such that each set contains classifiers in $\Ca$ with some $t \in \T $ (or with $\f{t}$, where $t \in \T$). Thus, for each $t \in \T$, Fuzzy Extraction produces (1) an orbit 
containing $t$ and elements in $\Ca$ with $t$, and, (2) an orbit 
consisting of elements in $\Ca$ with $\f{t}$ (but not necessarily containing $\f{t}$).

\paragraph*{Implications of Fuzzy Extraction Algorithm} 
The Fuzzy Extraction algorithm teases apart whether any two balanced accepted classifiers belong to the $\delta$-neighborhood of the same or different elements of $\T$. Note that,  $\WOra$ provides relatively vague information -- there could be a large number of valid pairs on which $\WOra$ behaves arbitrarily.  In the full paper, we establish that despite such imprecise information, Fuzzy Extraction  distinguishes $\delta$-neighborhoods of different elements of $\T$ successfully, and recovers all balanced accepted members of the neighborhoods.

\IncMargin{1em}
 \begin{algorithm}
\SetAlgoLined
\SetKwInOut{Input}{input}\SetKwInOut{Output}{output}
\Input{The universe $\X$ and an oracle $\WOra$ .}
\Output{ A collection of $2|\T|$ disjoint subsets 
of $\zoX$.}
\BlankLine
{
\ Find $\bV = \{c\in \zoX \, \mid \, \exists \Aw : ~ \WOra(\Aw,c)=1\}.$ \ Construct $\bVB = \bV \cap \B$.
Set $\C=1$}\;
 \While{$\bVB \neq \phi$}{
 Choose a classifier $\P_{\C}$ from $\bm{V}_{\B}$ and set $\bW_\C:=\{\P_{\C}\}$. If $\f{\P_{\C}} \in \bVB$, set $\bW_{\C}' = \{\f{\P_{\C}}\}$, else set $\bW_{\C}'= \phi$\;
\For{$c \in \bm{V}_{\B}\backslash \{ \P_{\C}\}$}{
 \If{$\CSa{c_{\C}}{c}=$``\,\normalfont{Situation A holds''}}{
 \eIf{$\Gzero{c_{\C}}{c} > \sqrt{2\delta}\X \geq \Gone{c_{\C}}{c}$}{
 update
 \[\bW_{\C}=\bW_{\C}\cup \{c\},  \, \, \bV_{\B} = \bV_{\B} \backslash \{c\},  \]
and if $\f{c} \in \bV_B$, update \[\bW_{\C}' = \bW_{\C}' \cup \{\f{c}\},  \bV_{\B} = \bV_{\B} \backslash \{\f{c} \} ;\]
}
 { update 
 \[\bW_{\C}'=\bW_{\C}' \cup \{c\}, \,\, \bV_{\B} = \bV_{\B} \backslash \{c\}, \]
and if $\f{c} \in \bVB$, update  \[\bW_{\C}=\bW_{\C}\cup\{\f{c}\},  \bV_{\B} = \bV_{\B} \backslash \{ \f{c}\}.\] \
 } 
 }
 }
  Set  $\bm{V}_{\B}=\bm{V}_{\B} \backslash\{\P_{\C} \}$, $\C=\C+1$\;}
 Return $\bW_1,\hdots, \bW_{\ell-1},\bW'_{1},\hdots, \bW'_{\ell-1}$, and additional sets $\bW_{\ell} = \{(c_i=1, \forall i \in \X) \}$ $\bW'_{\ell} = \{(c_i=0, \forall i \in \X) \}$, whenever a constant classifier is in $\bV$.  
 \caption{Fuzzy Extraction}\label{algo:fuzzy}
\end{algorithm}

\IncMargin{1em}
 \begin{algorithm}\label{algo:checksita}
\SetAlgoLined
\SetKwInOut{Input}{input}\SetKwInOut{Output}{output}
\Input{An ordered pair of classifiers $(c,u) \in \zoX$.}
\Output{Either
``Situation A holds" or
``Situation A does not hold".}
\BlankLine
\eIf{
      $\min \{|\Gzero{c}{u}|, |\Gone{c}{u} |\} \leq \sqrt{2 \delta } \X  < \max \{|\Gzero{c}{u}|, |\Gone{c}{u} |\}$ \, \\ {\normalfont and} \ \\
        $ \min \{|\Gdui{c}{u}|, |\Gtre{c}{u} |\} \leq \sqrt{2 \delta }\X  < \max \{|\Gdui{c}{u}|, |\Gtre{c}{u} |\} $ \
    }{return ``Situation A holds"}{``Situation A does not hold"}
 \caption{$\mathrm{CheckSituationA}$: Checks whether Situation A from Lemma \ref{lemma:distinguish} holds}
 \end{algorithm}
 
 \paragraph*{Intuition for the Sharp Extraction Algorithm}
We now focus on the Sharp Extraction algorithm.
Fix a strong oracle $\Ora$ with associated set $\T$ of classifiers, and let $\WOra$ be a $\delta$-weak relaxation of~$\Ora$.  Run Algorithm~\ref{algo:fuzzy} with weak oracle $\WOra$ to obtain a collection of orbits. 

For this informal discussion, fix $t \in \T$ and let $\bW$ be the orbit from Algorithm \ref{algo:fuzzy} that contains $t$, possibly together with some classifiers in $\Ca$ with $t$.
Algorithm~\ref{algo:sharp} applies a screening procedure to the elements of each orbit.  Applied to the members of $\bW$, the procedure may screen out some $u \in\bW$, meaning that it determines definitively that $u \notin \T$, but other vectors $v\in \bW$ may remain; in particular, $t$ will remain.  

The screening procedure invokes a primitive~$\MR$ (Definition \ref{def:MR}) with the key property that $\forall v \in \bW,~\MR(t,v) \in \bW$.  Thus, to test if $v\in \bW$ is a valid candidate for $t$, the algorithm tests whether $\MR(u,v) \in \bW$ for all $u\in \bW$.

\IncMargin{1em}
 \begin{algorithm}[H]
\SetAlgoLined
\SetKwInOut{Input}{input}\SetKwInOut{Output}{output}
\Input{ The universe $\X$ and the 
$\WO$ considered in Theorem \ref{thm:sharpthmsumm}. }
\Output{A list $P_1,\hdots, P_{m}, Q_1,\hdots, Q_{m}$ of classifiers of the universe $\X$.  }
\BlankLine
\emph{Run Fuzzy Extraction with the inputs $\X$ and $\WOra$. Let $\bW_1,\hdots, \bW_{m}, \bW_1',\hdots, \bW_m'$ denote the output. For $j=1,\hdots,m$, define $\Pi_j, \Pi_j', \Gamma_j, \Gamma_j' = \phi$. }\;
\For{$j=1,\hdots, m$}{\For{$\P \in \bW_j$}{if for all $\P' \in \bW_j, \MR(\P,\P') \in \bW_j$, update $\Pi_{j}  = \Pi_{j} \cup \P$ \; 
if for all $\P' \in \bW_j, \ML(\P,\P') \in \bW_j$, update $\Gamma_{j}  = \Gamma_{j} \cup \P$ \;} 
\For{$\P' \in \bW'_j$}{if for all $\P'' \in \bW'_j, \ML(\P',\P'') \in \bW'_j$, update $\Pi'_{j}  = \Pi'_{j} \cup \P'$ \; 
if for all $\P'' \in \bW'_j, \MR(\P',\P'') \in \bW'_j$, update $\Gamma'_{j}  = \Gamma'_{j} \cup \P'$\; }
}
\For{$j=1,\hdots,m$}{set $u_j=\MR(\Pi_j), v_j = \ML(\Pi'_j), w_j = \ML(\Gamma_j), x_j= \MR(\Gamma_j')$, and define

{\small\[P_j = \twopartdef{u_j}{|\ls{u_j}| \geq |\rs{v_j}|}{\f{v_j}}{|\rs{v_j}| > |\ls{u_j}|}, \qquad Q_j = \twopartdef{w_j}{|\rs{w}_j| \geq |\ls{x}_j|}{\f{x_j}}{|\ls{x}_j| > |\rs{w_j}|}.    \] }
}


Return $P_1,\hdots, P_m, Q_1,\hdots, Q_m $\;\tcp{The classifier $c$ with $\ls{P_j} \subseteq \ls{c}$ and $\rs{Q_j} \subseteq \rs{c}$ accurately recovers $\ls{t}, \rs{t}$ (or $\ls{[\f{t}]},\rs{[\f{t}]}$) upto a small error margin (Theorem \ref{thm:sharpthmsumm}).} 
  \caption{Sharp Extraction}\label{algo:sharp}
\end{algorithm}

Let $\In \subseteq \bW$ be the subset of $\bW$ screened in. Let us arbitrarily name these elements $u_1, u_2, \dots, u_k$. If there is exactly one element in $\In$, then $u_1 = t$.
This is excellent: the algorithm found an element of $\T$.  
Consider now the more general case of multiple elements.  Choose any ordering of $\In$, say, $(u_1, u_2,\dots, u_k)$.  Then since $u_1 \in \In$, we have $w=\MR(u_1, u_2) \in \bW$. Now, since $u_3 \in \In$, $\MR(w,u_3) \in \bW$.  We can continue in this way until we have merged in $u_k$, and we see by induction that at every step the merge remains in $\bW$. 

$\MR$ is commutative and associative, so we can assume without loss of generality that $u_1 = t$, which will simplify the description of the properties of the merging of all the classifiers in~$\In$.
  We first define the operation.
\begin{definition}[$\MR$] \label{def:MR}
For a set of classifiers ${\cal C}=\{\P_1,\hdots, \P_k\}$, define the {\em $\MR$} operation applied to ${\cal C}$ by $\MR(\{\P_1,\hdots, \P_k\})$ to be a new classifier $w$ such that
\begin{equation}\label{eq:mergeright}
\rs{w}=\cup_{j \in [k ]} \rs{\P_j}, \qquad \ls{w}=\X \backslash \rs{w}
\end{equation}
{\em $\ML$ } is defined analogously and denoted $\ML(\{\P_1,\hdots, \P_k\})$.
\end{definition}

Let $w=\MR(t,u)$ for an arbitrary $u\in\zoX$. Then $\rs{w}$ contains all the elements with positive classification under $t$ (and other elements that are positive under $u$).
Since $\ls{w}$ contains none of these, we have that $\ls{w} \subseteq \ls{t}$.

Let $w=\MR(\{u \in \In\})$. Then by the above reasoning also for this $w$, we have $\ls{w} \subseteq \ls{t}$.  To argue that $\ls{w}$ recovers most of $\ls{t}$, we show that $\ls{t} \cap \rs{w}$ must be small.  This follows from the fact that $w \in \bW$, as argued above. 
This ensures that $\ls{t} \cap \ls{w}$, is in fact, large. 
A similar reasoning applies to the $\ML{}$ procedure, which is used to create another subset $\In' \subseteq \bW$ with the property that $\rs{\ML(\In')}$ has a large intersection with $\rs{t}$.

In its final phase, still focusing on a single $t$ and its corresponding orbit $\bW$, the algorithm combines the left side of the output of the $\MR{}$ procedure and the right side of the output of the $\ML{}$ procedure to create a classifier that substantially agrees with $t$ (for elements $i \notin \ls{(\MR(\In))}\cup \rs{(\ML(\In'))}$ it assigns an arbitrary value).
This completes the high level intuition for the Sharp Extraction algorithm. Theorem \ref{thm:sharpthmsumm} provides the formal guarantees.

\section{Discussion on Assumptions}\label{sec:assumptions}
Our main theorem relies on the following crucial assumption regarding the structure of $\T$.

  \begin{assumption}\label{assm:summary}
Assume that every $t \in \T$ obeys the following structure.
\begin{enumerate}
    \item Every $t \in \T$ that is not a constant classifer, must be $4 \delta$-balanced. 
    \item 
   For any $t, u \in \T$, if $\f{t}\neq u$, then at most one of $\Gzero{t}{u}, \Gone{t}{u}, \Gdui{t}{u} $ and $ \Gtre{t}{u}$ can be empty. Furthermore, whenever one of these sets is non-empty, it must  contain strictly more than  $2\sqrt{2 \delta} \X$ elements (Figure \ref{fig:faithfulness}).
   \item For every $t \in \T$ such that $\f{t} \notin \T$, let $\WOra(\Aw,c) = 0$ whenever $c$ is in $\Ca$ with $\f{t}$. 
\end{enumerate}
\end{assumption}

We provide some intuition for the assumptions, deferring details to the full paper. 
Note that, if $t \in \T$ is imbalanced, then most $c \in \Gamma_\delta(t)$ will also be imbalanced. However, 
imbalanced classifiers are problematic: they belong to $\delta$-neighborhoods of more than one $t \in \T$, even when these are far apart. The presence of several imbalanced classifiers confuses our algorithms.
By requiring that all $t \in \T$ be well balanced ( Assumption~\ref{assm:summary}(1)), we ensure that sufficiently many $c \in \Gamma_\delta(t)$ are also balanced. 

To recover $\T$, our algorithms need to tease apart the following situations for any two classifiers $u,v \in \zoX$: (a) $\exists t \in \T $ such that $u,v \in \Gamma_\delta(t)$, and (b) $\nexists t \in \T$ such that $u,v \in \Gamma_{\delta}(t)$. Our intuition is that if elements of $\T$ are well-separated as defined via transportation costs, then this should be possible; however, our proof requires Assumption~\ref{assm:summary}(2) 
that implies separation but is not equivalent. We do not know if this stronger condition can be relaxed.

Finally, for any $t \in \T$, the flipped classifier $\f{t}$ may or may not belong to $\T$. But, the neighborhoods of $t$ and $\f{t}$ are identical---this creates additional challenges when $t \in \T$ but $\f{t} \notin \T$. Assumption \ref{assm:summary}(3) protects from complications arising in this case.

\section*{Acknowledgements}
This research was conducted, in part, while the authors were at Microsoft Research, Silicon Valley. C.D.~was supported in part by  NSF grant CCF-1763665 and Microsoft Research. C.I.~was supported by Microsoft Research and the Sloan Foundation. G.R.~ received funding from the European Research Council (ERC) under the European Union’s Horizon 2020 research and innovation programme (grant agreement No. 819702), the Israel Science Foundation (grant number 5219/17), and Microsoft Research. P.S.~was supported by the Center for Research on Computation and Society, Harvard University, and in part by Microsoft Research.

\bibliographystyle{plain}

\bibliography{proposal,salil,references}

\end{document}